\documentclass[10pt,twocolumn,letterpaper]{article}

\usepackage{cvpr}
\usepackage{times}
\usepackage{epsfig}
\usepackage{graphicx}
\usepackage{amsmath}
\usepackage{amssymb}
\usepackage{subcaption}

\usepackage{multirow}

\usepackage[pagebackref=true,breaklinks=true,letterpaper=true,colorlinks,bookmarks=false]{hyperref}

 \cvprfinalcopy 

\def\cvprPaperID{5025} 

\ifcvprfinal\fi
\begin{document}

\title{Learning Spatial Pyramid Attentive Pooling in Image Synthesis and Image-to-Image Translation}

\author{Wei Sun and Tianfu Wu\\
Department of ECE and the Visual Narrative Initiative, 
North Carolina State University\\
{\tt\small \{wsun12, tianfu\_wu\}@ncsu.edu}
}

\maketitle

\begin{abstract}
  Image synthesis and image-to-image translation are two important generative learning tasks. Remarkable progress has been made by learning Generative Adversarial Networks (GANs)~\cite{goodfellow2014generative} and cycle-consistent GANs (CycleGANs)~\cite{zhu2017unpaired} respectively. This paper presents a method of learning Spatial Pyramid Attentive Pooling (SPAP) which is a novel architectural unit and can be easily integrated into both  generators and discriminators in GANs and CycleGANs. The proposed SPAP integrates Atrous spatial pyramid~\cite{chen2018deeplab}, a proposed cascade attention mechanism and residual connections~\cite{he2016deep}. It leverages the advantages of the three components to facilitate effective end-to-end generative learning: (i) the capability of fusing multi-scale information by ASPP; (ii) the capability of capturing relative importance between both spatial locations (especially multi-scale context) or feature channels by attention; (iii) the capability of preserving information and enhancing optimization feasibility by residual connections. Coarse-to-fine and fine-to-coarse SPAP are studied and intriguing attention maps are observed in both tasks. In experiments, the proposed SPAP is tested in GANs on the Celeba-HQ-128 dataset~\cite{karras2017progressive}, and tested in CycleGANs on the Image-to-Image translation datasets including the Cityscape dataset~\cite{cordts2016cityscapes}, Facade and Aerial Maps dataset~\cite{zhu2017unpaired}, both obtaining better performance.    
\end{abstract}

\section{Introduction}
 
Image synthesis is an important task in computer vision  which aims at synthesizing realistic and novel images by learning high-dimensional data distributions. Image-to-Image translation is usually built on image synthesis, especially unsupervised translation (i.e., translation with unpaired images). It provides a general framework for many computer vision tasks such as super-resolution~\cite{ledig2017photo, shi2016real}, image colorization~\cite{isola2017image}, image style generation~\cite{zhu2017unpaired} and image segmentation~\cite{sankaranarayanan2018learning}. 
Generative adversarial networks (GANs)~\cite{goodfellow2014generative} are one of the main methods for image synthesis. GANs utilize a two-player game formulation in which one player, \emph{the generator} learns to synthesizes images that are indistinguishable from the training data, while the other player, \emph{the discriminator} is trained to differentiate between the generated images and real ones. The generator and discriminator are trained simultaneously by solving a notoriously hard adversarial loss minimax problem. Built on GANs, cycle-consistence GANs (CycleGANs)~\cite{zhu2017unpaired} are one of the main approaches for image-to-image translation. CycleGANs introduce a cycle-consistent loss term into GANs. 

\begin{figure}[t]
    \centering
    \includegraphics[width = 1.0\linewidth]{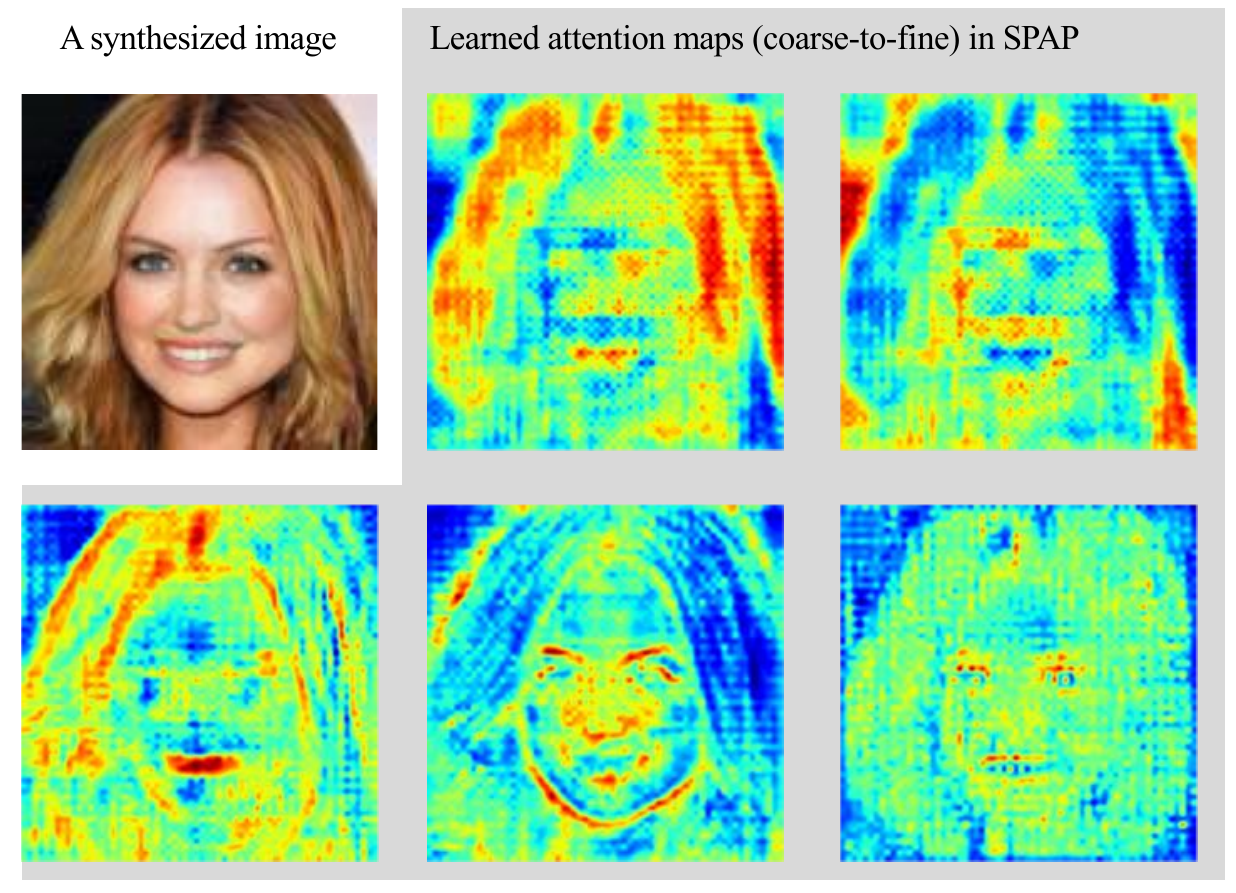}
    \caption{Illustration of the proposed Spatial Pyramid Attentive Pooling (SPAP) in image synthesis using GANs~\cite{goodfellow2014generative} on the Celeba-HQ-128 dataset~\cite{karras2017progressive}. The attention maps, visualized usng heat maps, are learned in the Atrous spatial pyramid from coarse levels to fine levels (i.e., the dilation rates go from large to small) to fuse different levels.  See text for details. (Best viewed in color)}
    \label{fig:introduction}
\end{figure}

Due to the difficulty of solving the minimax problem in practice, many efforts have been devoted to improve the stability of training, the quality of generated images and the capability of generating high-resolution images. Most efforts are mainly focused on loss function design~\cite{salimans2018improving, arjovsky2017wasserstein, metz2016unrolled, che2016mode, zhao2016energy, nowozin2016f}, regularization and normalization schema~\cite{gulrajani2017improved, miyato2018spectral}, introducing heuristic tricks \cite{salimans2016improved, odena2016conditional, heusel2017gans}, and training protocol such as the progressively trained GANs~\cite{karras2017progressive}. Less attention has been paid to neural architecture design, especially for the generators. The intuitive idea of this paper is that exploring new architectures could improve performance of generative learning in a way complementary to existing efforts. 
The goal of this paper is to design a generic and light-weight architectural unit that can be easily integrated into both generators and discriminators of GANs and CycleGANs.

The paper presents a spatial pyramid attentive pooling (SPAP) building block which can be used to substitute a existing layer in the generator a GAN or a CycleGAN. It is built on three ubiquitous components in network architecture design and integrates them in a novel way for generic applicability. 
\begin{itemize}
    \item Atrous convolution or dilated convolutions~\cite{chen2014semantic}. The objective is to capture multi-scale information in the input feature map by computing an Atrous spatial pyramid~\cite{chen2018deeplab}. How to aggregate information in the Atrous spatial pyramid is usually a task-specific problem. For example, in semantic segmentation tasks, they are concatenated together, followed by a $1\times 1$ convolution, so called ASPP to fuse the information in one of the state-of-the-art methods, DeepLab~\cite{chen2018deeplab}. The ASPP has not been studied in generative learning. However, we observed that this vanilla aggregation is not good enough for generative learning tasks. So, we propose to integrate attention mechanism in the pyramid.     
    \item Self-attention mechanism~\cite{zhang2018self}. It has been recently studied in GANs with significant performance improvement~\cite{zhang2018self}. It modulates response at a position as a weighted sum of features at all positions with learned weights (i.e., a simple position-wise fully connected layer). It is not straightforward to extend this type of self-attention to aggregate multi-scale information. We proposed a novel cascade based scheme to integrate information between successive levels in the pyramid, either coarse-to-fine or fine-to-coarse. Fig.~\ref{fig:introduction} shows an example of synthesized face and the corresponding learned coarse-to-fine attention maps. The proposed cascade attention scheme implicitly implements the progressive training ideas~\cite{karras2017progressive}. We show in our experiments that the proposed cascade attention is more effective than the vanilla ASPP method~\cite{chen2018deeplab}.  
    \item Residual connections~\cite{he2016deep}. We adopt a convex combination between the attention modulated information and the original input information, as done in~\cite{zhang2018self}. The weight is learned. This residual connection will help both exploit original information and keep the feasibility of optimization.    
\end{itemize}

In summary, the propose SPAP building block harnesses the advantages of the above stated components in generative learning tasks. In experiments, the proposed method is tested in both image synthesis tasks using the state-of-the-art SNDCGANs~\cite{kurach2018gan}, and unpaired image-to-image translation tasks using the popular CycleGANs~\cite{zhu2017unpaired}. We obtain significantly better performance than the vanilla SNDCGANs and CycleGANs and the baseline ASPP~\cite{chen2018deeplab} module. Although our models are much smaller, we obtain comparable performance to the most recent extension of CycleGANs, the SCAN~\cite{yang2016stacked} which use stacked CycleGANs in the progressive training protocol.  

\section{Related Work and Our Contributions}
We first briefly overview GANs based generative learning and the related applications of Atrous convolution and attention mechanism. 

\textbf{Generative Adversarial Networks}
Generative Adversarial Networks (GANs) have achieved great success in various image generation tasks, including image-to-image translation \cite{isola2017image, zhu2017unpaired, taigman2016unsupervised, liu2016coupled, huang2018multimodal}, image super-resolution \cite{ledig2017photo, sonderby2016amortised} and text-to-image synthesis \cite{reed2016generative, reed2016learning, yang2016stacked}. Despite the success, the training of GANs is notorious to be unstable and sensitive to the choices of hyper-parameters. Several directions have attempted to stabilize the training of GAN and improve the generated sample diversity, including designing new network architectures \cite{radford2015unsupervised, zhang2017stackgan, karras2017progressive}, modifying the learning objectives and dynamics \cite{salimans2018improving, arjovsky2017wasserstein, metz2016unrolled, che2016mode, zhao2016energy, nowozin2016f}, adding regularization methods \cite{gulrajani2017improved, miyato2018spectral} and introducing heuristic tricks \cite{salimans2016improved, odena2016conditional, heusel2017gans}. Recently spectral normalized model together with projection-based discriminator  \cite{miyato2018cgans} and \cite{brock2018large} greatly improves class-conditional image generation on ImageNet.

\begin{figure*} [t]
    \centering
    \includegraphics[width = 1.0\linewidth]{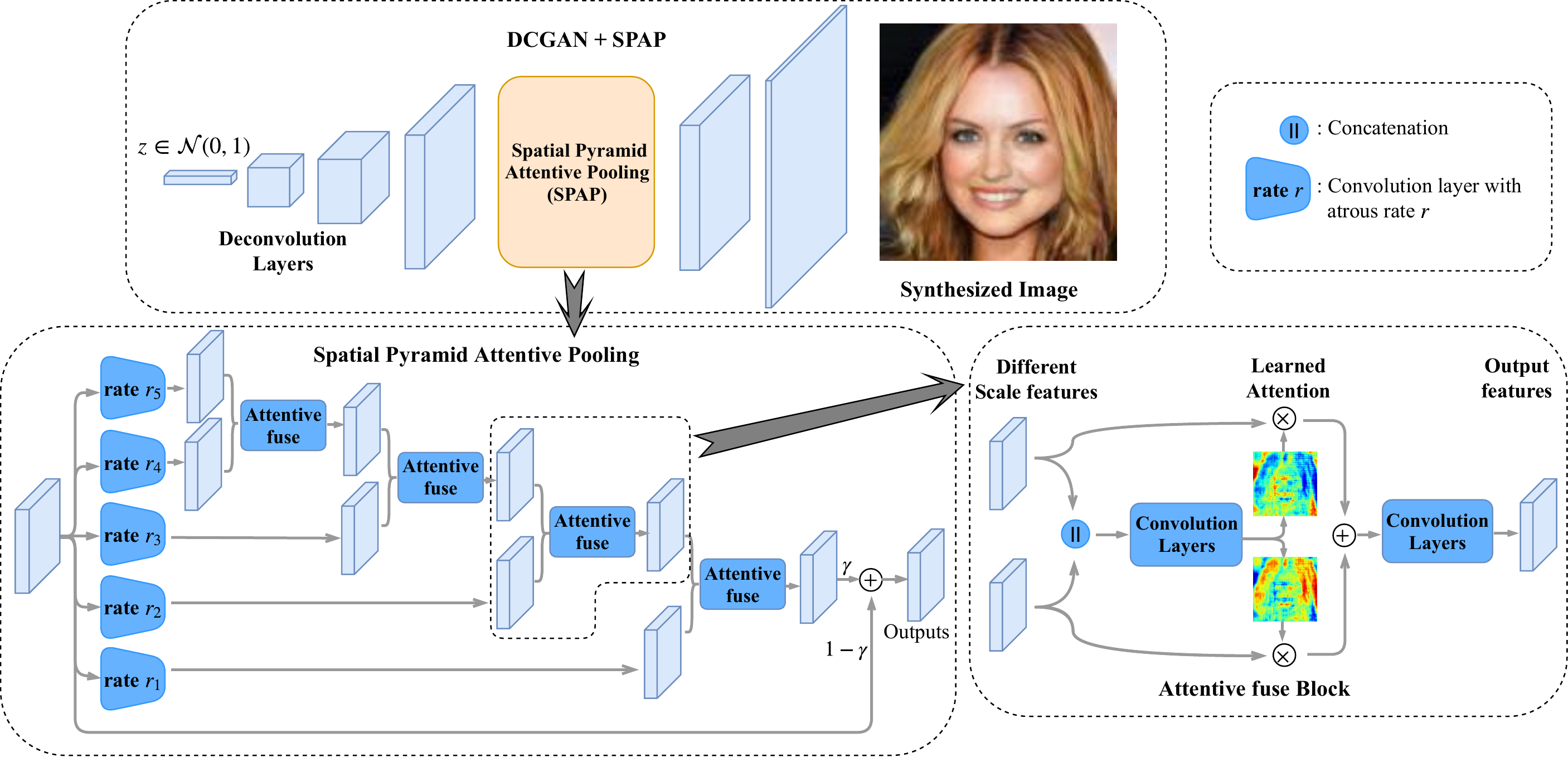}
    \caption{The proposed Spatial Pyramid Attentive Pooling (SPAP) buidling block. \textit{Top:} Illustration of the integration of the proposed SPAP in the generator of an unconditional GAN for image synthesis. \textit{Bottom-left:} The detailed neural architecture of the proposed SPAP. 
    \textit{Bottom-right:} The operation of the attentive fuse component between two consecutive levels in the pyramid.}
    \label{fig:ASPP}
\end{figure*}

\textbf{Atrous Convolution}
Atrous convolution is first introduced in \cite{chen2014semantic} to increasing receptive field while keeping the feature map resolution unchanged. 
Atrous Convolution (Dilated Convolution) which can effectively incorporate surrounding context by enlarging receptive field size of kernels, has been explored in image segmentation \cite{chen2018deeplab, chen2017rethinking, chen2018encoder} and object detection \cite{papandreou2015modeling, dai2016r, huang2017speed, li2018detnet}. Atrous Spatial Pyramid pooling (ASPP) \cite{chen2018encoder, chen2017rethinking, chen2018deeplab}, which exploits the multi-scale information by applying several parallel atrous convolution with different rates, has prove performance on segmentation tasks and promising results on several segmentation benchmarks. DenseASPP \cite{yang2018denseaspp} connects a set of atrous convolution layers in a dense way, which effectively generates features that covers a large range. In \cite{wei2018revisiting}, multiple dilated convolutional blocks of different rates are applied for dense object localization and then weakly supervised semantic segmentation. In this paper, we adopt multi atrous convolution to increase the ability of Generator $\mathbb{G}$ and Discriminator $\mathbb{D}$.

\textbf{Attention Models}
Recently attention mechanisms has been exploited in many tasks, including VQA \cite{yang2016stacked} and image classification \cite{zhu2017learning, woo2018cbam}. SA-GAN \cite{zhang2018self} first exploit attention to GAN by adding self-attention block to Generator and Discriminator to improve the ability of network to model global structure. In this work, we apply spatial attention when fusing feature maps from different atrous convolution layers in generator $\mathbb{G}$.

\textbf{Our Contributions.} This paper makes the following two main contributions to the field of generative learning. 
\begin{itemize}
    \item It presents a novel architectural building block, namely spatial pyramid attentive pooling (SPAP), which integrates Atrous spatial pyramid for capturing multi-scale information, a cascade attention scheme for fusing information between multi-scale levels in the pyramid, and a residual connection for feature reuse and enhancing optimization feasibility. To our knowledge, it is the first work that investigates the effects, for generative learning tasks, of Atrous spatial pyramid and the cascade attentive fusion of information from different levels in the pyramid. 
    \item It shows better performance in a series of image synthesis tasks and image-to-image translation tasks.
\end{itemize}

\section{The Proposed Method}
In this section, we first briefly introduce the background of GANs and CycleGANs to be self-contained. Then, we present the proposed SPAP module. 
\subsection{Background}
\textbf{Unconditional GAN} A GAN consists of a generator ($\mathbb{G}$) that map random noise $z$ to samples and a discriminator ($\mathbb{D}$) that distinguishes the generated samples from the real samples. For unconditional GANs, the basic framework can be viewed as a two-player game between $\mathbb{G}$ and $\mathbb{D}$, and the objective is to find a Nash equilibrim to the min-max problem,
\begin{equation}
    \min_{G} \max_{D} \mathbf{E}_{x \sim q(x)}[\log D(x)] + \mathbf{E}_{z\sim p(z)} [\log(1-D(G(z)))]
\end{equation}
where $z \in \mathbb{R}^{d_z}$ is a latent variable from distributions such as $\mathcal{N}(0, 1)$ or $\mathcal{U} [-1, 1]$. For generation model on images, deconvolution and convolution neural networks are usually utilized in Generator $\mathbb{G}$ and Discriminator $\mathbb{D}$ respectively.

\textbf{Image-to-Image translation with CycleGAN} GANs have shown improved results in image to image translation tasks \cite{isola2017image, yi2017dualgan}. Recent work by Zhu. \textit{et al} \cite{zhu2017unpaired} has tackled the unpaired image-to-image translation task with a combination of adversarial and cycle-consistency losses. Let's consider two domain ${X}$ and ${Y}$, the CycleGAN model contains two translator model $\mathbb{G}$ and $\mathbb{F}$ which map ${X} \rightarrow {Y}$ and ${Y} \rightarrow {X}$ respectively. The model has two additional adversarial discriminators $\mathbb{D_X}$ and $\mathbb{D_Y}$ aiming to between real images and translated images in each domain. The loss function of translator and discriminator can be expressed as

\begin{equation}
    \begin{aligned}
    \mathcal{L}_{GAN}(G, D_X, X, Y) & = \mathbf{E}_{y \sim p_{data}(y)}[\log D_Y(y)] \\
    & + \mathbf{E}_{x \sim p_{data}(x)}[\log(1 - D_Y(G(x))] \\
    \end{aligned}
\end{equation}

\begin{equation}
    \begin{aligned}
    \mathcal{L}_{cyc}(G,F) & = \mathbf{E}_{x \sim p_{data}(x)}[||F(G(x)) - x||_1] \\
    & + \mathbf{E}_{y \sim p_{data}(y)} [||G(F(y)) - y||_1] \\
    \end{aligned}
\end{equation}

\noindent and full objective of CycleGAN is:
\begin{equation}
    \begin{aligned}
    \min_{G,F} \max_{D_X,D_Y} \mathcal{L}(G,F,D_X,D_Y) &= \mathcal{L}_{GAN}(G, D_Y,X,Y) \\
    & + \mathcal{L}_{GAN}(F, D_X, Y, X) \\
    & + \lambda \mathcal{L}_{cyc}(G,F)
    \end{aligned}
\end{equation}

\subsection{The Proposed SPAP}

As we briefly discussed in the introduction, our method is motivated by Atrous Spatial Pyramid Pooling (ASPP)~\cite{chen2018deeplab} where parallel atrous convolution layers with different rates capture multi-scale information. Most of GANs and related models for image generation and translation tasks are based on convolutional layers with small kernels ($3\times 3$ or $4\times 4$) in order to keep both computation and the number of parameters contained. Small kernel convolution process information in a neighborhood. Multi-scale information captured by atrous convolution should also improve performance of image synthesis with GANs.

Figure \ref{fig:ASPP} illustrates the structure of our SPAP module and its integration in an unconditional GAN structure. In the intermediate feature of the model, we apply several parallel convolution layers with different atrous rate, of which each capture different scale information. In ASPP module of image segmentation model, the features extracted are further processed and fused by channel concatenation and $1\times 1$ convolution. Image synthesis task is different from segmentation, and we know that different regions in the image should focus on different scale information. Spatial attention is not included in $1\times 1$ convolution. In our module, we design an attentive fuse component to  two feature maps of different scale each time. For the attentive fuse layer, as shown in top right of Fig.~\ref{fig:ASPP}, giving feature maps of two convolution layers $f_i$ and $f_{i+1}$, we introduce a spatial attention layer to learn a dynamic combination of the two features, 
\begin{equation}
    AttenFuse(f_i, f_{i+1}) = f_i \odot \alpha_i + f_{i+1} \odot (1-\alpha_{i+1})
\end{equation}
where $\alpha_i = Atten(f_i, f_{i+1})$ indicates the pixel-wise attention map predicted by sequence of convolutional neural network followed by sigmoid activation, and $\odot$ means element-wise product. Then fused multi-scale feature is added to the input feature map by a learnable scale parameter $\gamma$, so the final output would be,
\begin{equation}
    y_{out} = \gamma * o + (1 - \gamma) * x_{in}
\end{equation}

For the order of fusing different scale features, we experiment with various ways. For larger dilate convolution rate, the results are coarse. So we experiment with direction of coarse-to-fine as well as fine-to-coarse.

\begin{figure}[!h]
    \centering
    \includegraphics[width = 0.7\linewidth]{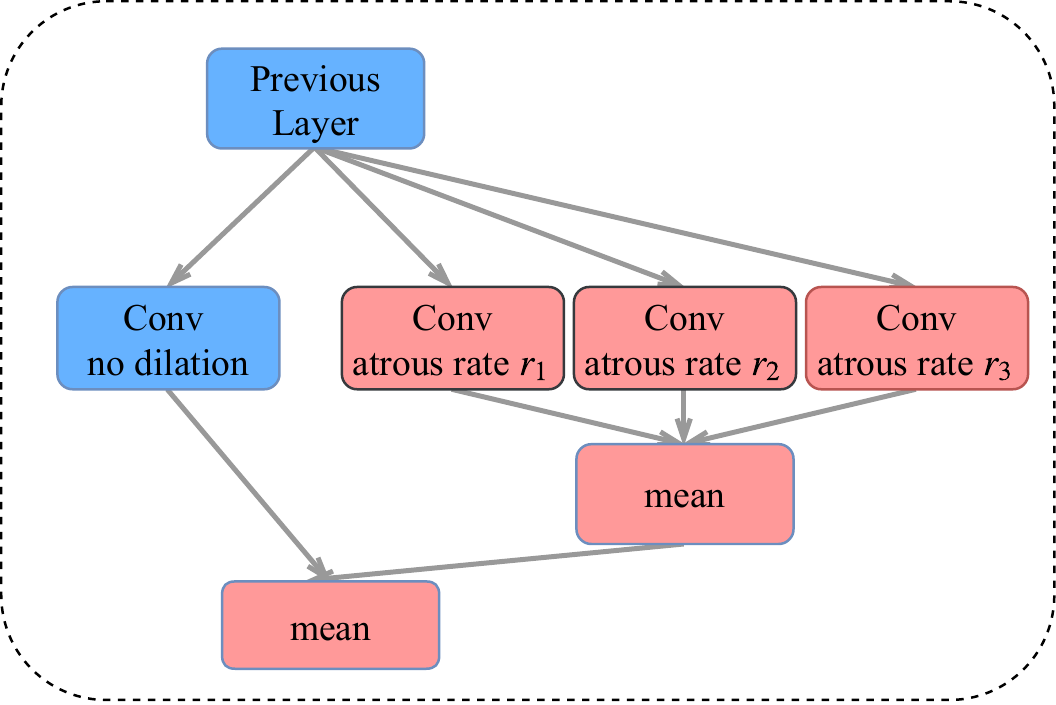}
    \caption{Adding atrous convolution to discriminator}
    \label{fig:AtrousDis}
 \end{figure}

\noindent\textbf{Discriminator.} 
The proposed SPAP module can be used in discriminator too. However, we observed that it is more important to improve the expressive power of generators in GANs and the job for discriminator is relatively simpler (i.e., binary classification between real vs fake). In addition, it will be clearer to show the capability of the proposed SPAP module if we use it only in generators, as well as for simplicity. That being said, we use much simpler aggregation scheme for discriminator. We only utilize atrous convolution since it can also increase the receptive field and improve the discriminative capability. Vanilla ASPP module used in segmentation network will introduce more parameter and make $\mathbb{D}$ harder to train. Following the idea of \cite{wei2018revisiting}, we add several parallel dilation convolutions to discriminator in a simple way. As shown in Fig.~\ref{fig:AtrousDis}, inside blue rounded rectangle are layers included in the original discriminator, and red ones are added atrous convolution layers. For a specific convolution layer, we can add several parallel convolutions with various atrous rates, then these layers are included in the whole network by calculating mean of outputs, and finally average with the origin no-atrous convolution layer. With this structure, we can effectively increase the receptive field of convolutional network without introducing too many parameters. 

\section{Experiments}

We test the proposed SPAP module in a series of generative learning tasks. 

\subsection{Datasets}
\textbf{CELEBA-HQ} CelebA-HQ is a high quality version of CelebA dataset, which consists of 30000 of the images in 1024x1024 resolution \cite{karras2017progressive}. We use the 128x128x3 version obtained by the code provided by the author\footnote{Code available at \url{https://github.com/tkarras/progressive_growing_of_gans
}}, randomly split 27000 images as training dataset and remaining 3000 image as testing dataset. 

\textbf{Image-to-Image Translation} To demonstrate the capability of our proposed SPAP module in GAN related networks, we test on unsupervised image-to-image translation problem with CycleGAN model \cite{zhu2017unpaired}. We conduct experiments on Cityscapes Labels $\Leftrightarrow$ Photo, Maps $\Leftrightarrow$ Aerial, Facades $\Leftrightarrow$ Labels and Horse $\Leftrightarrow$ Zebra datasets. We compare with CycleGAN and SCAN \cite{Li_2018_ECCV} and Pix2Pix \cite{isola2017image} results on 256x256 resolution images.

\subsection{Evaluation Metrics}
\textbf{Fr\'echet Inception distance (FID)} FID score was introduced in \cite{heusel2017gans}. Samples from $\mathcal{X}$ and $\mathcal{Y}$ are first embeded into a feature space by a specific later of InceptionNet. Both these feature distributions are modeled as multi-dimensional Gaussians parameterized by their respective mean and covariance. Then the Fr\'echet distance is measured by 
\[\textrm{FID} = ||\mu_x - \mu_y||^2 + \textrm{Tr}(\Sigma_x + \Sigma_y - 2(\Sigma_x\Sigma_y)^\frac{1}{2})\]
where ($\mu_x$, $\Sigma_x$) and ($\mu_y$, $\Sigma_y$) denote the mean and covariance of the real and generated image distributions respectively. FID score is used to evaluate generative models for no-labeled data and is robust to various manipulations and sensitive to mode dropping \cite{lucic2017gans}. 

\textbf{FCN Segmentation Score} For Cityscape Label $\Leftrightarrow$ Photo dataset, we apply segmentation scores to evaluates how interpretable the translated photos. An off-the-shelf FCN segmentation network \cite{long2015fully} is applied to the translated images for predicting labels, then three standard segmentation metrics is calculated against the ground truth labels, including per-pixel accuracy, the per-class accuracy and the mean class accuracy.

\textbf{PSNR and SSIM} For Facade $\Leftrightarrow$ Label and Maps $\Leftrightarrow$ Label dataset, we calculate PSNR and SSIM \cite{wang2004image} for quantitive evaluation. PSNR can measure color similarity and SSIM can measure structual similarity between translated images and ground truth.

\subsection{Network Structure and Implement Details}
For unconditional image generation network, we follow the SNDCGAN setting of  \cite{kurach2018gan}. Spectral norm is applied \cite{miyato2018spectral} to both generator and discriminator. For all models, we use the Adam optimizer \cite{kingma2014adam}, learning rate for discriminator and generator are 0.0002 and 0.0001. Batch size is set as 64 and total training steps are 100k. When deploying SPAP, we add one module  after the 64x64 size feature map of generator, and also add parallel atrous convolution layer in second downsample layer of discriminator as show in Fig.~ \ref{fig:AtrousDis}. 

For image-to-image translation task, we adopt the network structure setting of \cite{zhu2017unpaired}. For $256\times 256$ images, the generator network contains two stride-2 convolution, 9 residual blocks and two deconvolution layers with stride $\frac{1}{2}$. We conduct one SPAP module after the first upsample deconvolution layer, which has a feature map size of 128x128. For discriminator, three atrous convolutions are deployed to the third convolution layer, where the input feature map size is 64x64, to increase discriminative ability. Adam optimizer \cite{kingma2014adam} with $\beta_{1} = 0.5$ and $\beta_2 = 0.999$ is applied, learning rate set as 0.0002 for the first 100 epoch and linearly decay to zero in the next 100 epoches.

For SPAP module, we apply three 3$\times$3 atrous convolutions with different rates, one 3$\times$3 and then one 1$\times$1 convolution without atrous rate 1. For image generation on CelebA-HQ, we adopt rates (3, 5, 7) and for image-to-image translation task, a larger atrous rate (6, 12, 18) is applied where images resolution is higher. For both tasks, rates are set to (3, 5, 7) when deploying atrous convolution to discriminator $\mathbb{D}$. For all experiments, we start to update the parameters of SPAP and Atrous convolution after a number of training steps (40k steps for unconditional GAN and 100 epoch for CycleGAN model.)


\begin{figure*}
	\centering
	\captionsetup[subfigure]{labelformat=empty}
	\begin{subfigure}[b]{.33\linewidth}
	\includegraphics[width=\linewidth]{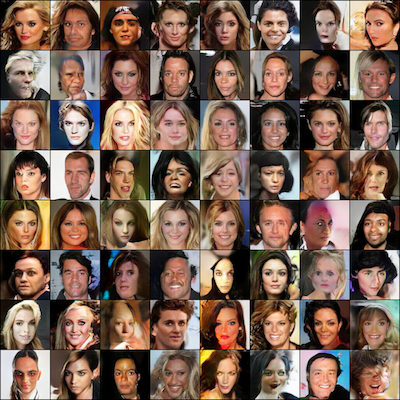}
	\caption{SPAP FID: 18.47}\label{fig:gull}
	\end{subfigure}
	\begin{subfigure}[b]{.33\linewidth}
	\includegraphics[width=\linewidth]{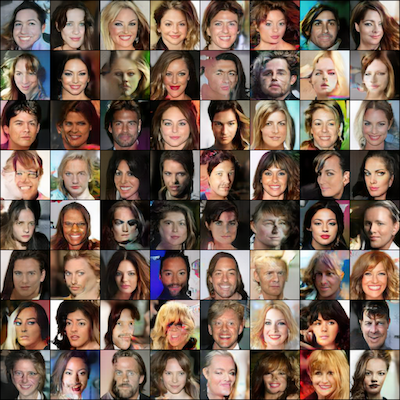}
	\caption{ASPP FID: 22.55}\label{fig:gull}
	\end{subfigure}
	\begin{subfigure}[b]{.33\linewidth}
	\includegraphics[width=\linewidth]{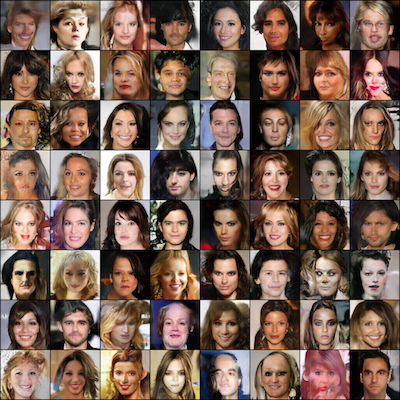}
	\caption{Vanilla FID: 25.42}\label{fig:gull}
	\end{subfigure}
	
	\caption{Examples of generated images by the proposed model trained on CelebA-HQ. }
	\label{fig:generated_faces}
\end{figure*}

\subsection{Experimental Results} 
\textbf{Unconditional GAN on CelebA-HQ} Sample of Generated images and its related FID score is show in Fig.~\ref{fig:generated_faces}. Table \ref{celebahq-fid} showes Fr\'echet Inception Distance (FID) with SNDCGAN + SPAP, compared with SNDCGAN \cite{kurach2018gan} and Self-Module \cite{chen2018self}. By just adding vanilla ASPP module to both generator and discriminator, FID score can be improved to 22.55, whiling applying SPAP in a coarse-to-fine order to generator and dilated convolutions to discriminator in SNDCGAN structure significantly improve FID score from 25.42 to 18.47. SPAP module in fine-to-coarse order also improve the FID socre to 19.73. This improvement demonstrate the effectiveness of the proposed SPAP in $\mathbb{G}$ and Atrous in $\mathbb{D}$ mechanism.

\begin{table}[!h]
	\caption{FID score on CELEBA-HQ (smaller is better)}\label{celebahq-fid}
	\centering
	\resizebox{0.48\textwidth}{!}{
	\begin{tabular}{cccc}
	\hline
	\textbf{Model} & \textbf{Parameters in $\mathbb{G}$} & \textbf{Best} &\textbf{Median} \\
	\hline
	Vanilla SNDCGAN &20.065M  & 25.42 & 26.11\\
	SELF-MOD \cite{chen2018self} &- & 22.51 &- \\
	SNDCGAN+ASPP &20.762M & 20.07 &21.26 \\
	SNDCGAN+SPAP (fine to coarse) &20.756 M & 19.73  &20.73\\
	SNDCGAN+SPAP (coarse to fine) &20.756 M & \textbf{18.47} & 20.11 \\
	\hline 
	\end{tabular}}
\end{table}

\begin{figure*}
\centering
\begin{tabular}{cccccc}

\includegraphics[width = 1.0in]{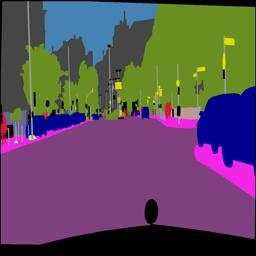} &
\includegraphics[width = 1.0in]{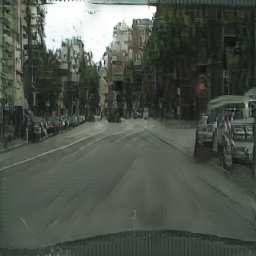} &
\includegraphics[width = 1.0in]{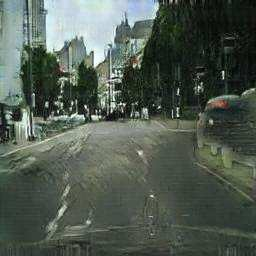} &
\includegraphics[width = 1.0in]{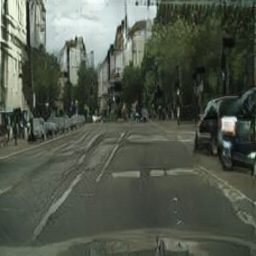} &
\includegraphics[width = 1.0in]{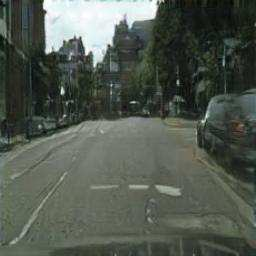} &
\includegraphics[width = 1.0in]{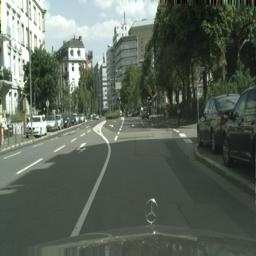}\\

\includegraphics[width = 1.0in]{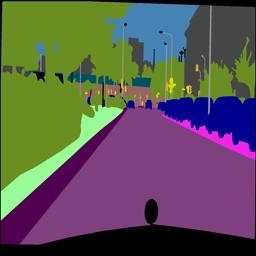} &
\includegraphics[width = 1.0in]{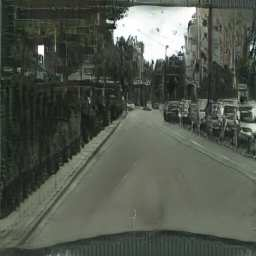} &
\includegraphics[width = 1.0in]{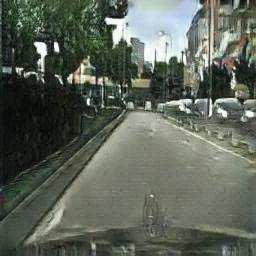} &
\includegraphics[width = 1.0in]{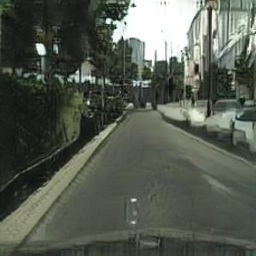} &
\includegraphics[width = 1.0in]{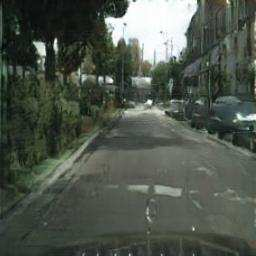} &
\includegraphics[width = 1.0in]{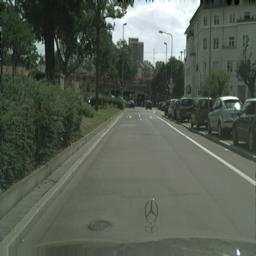}\\

\includegraphics[width = 1.0in]{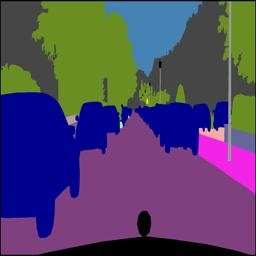} &
\includegraphics[width = 1.0in]{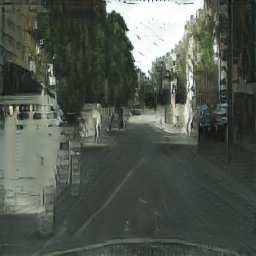} &
\includegraphics[width = 1.0in]{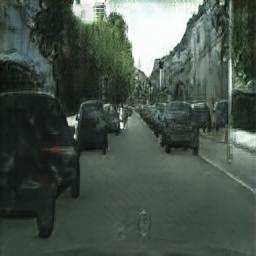} &
\includegraphics[width = 1.0in]{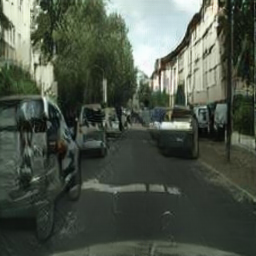} &
\includegraphics[width = 1.0in]{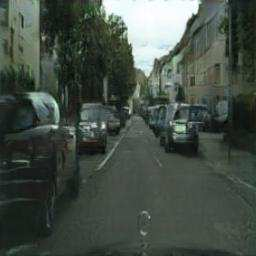} &
\includegraphics[width = 1.0in]{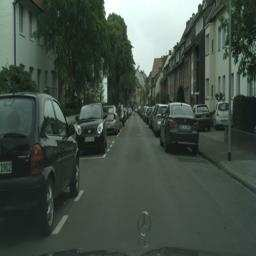}\\
Label & CycleGAN & SCAN  & Ours & Pix2Pix & Ground Truth
\\
\end{tabular}
\caption{Comparisons on Cityscapes dataset of 256x256 resolution. Images of CycleGAN and SCAN from \cite{Li_2018_ECCV}. Our results generated with SPAP in coarse-to-fine order.} \label{fig:cityscapes}
\end{figure*}

\begin{table*}[!h]
\caption{FCN segmentation scores on Cityscape Photo $\Leftrightarrow$ Label}  
\label{cityscape_fcn}
\centering
\begin{tabular}{ccccccc}
\hline
\multirow{2}{*}{Method} &\multicolumn{3}{c}{\textbf{Labels $\Rightarrow$ Photo}}  & \multicolumn{3}{c}{\textbf{Photo $\Rightarrow$ Label}}  \\
 & \textbf{Pixel acc.} &\textbf{Class acc.} &\textbf{Class IoU} &\textbf{Pixel acc.} &\textbf{Class acc.} &\textbf{Class IoU}\\
\hline
CycleGAN    &0.52  &0.17  &0.11  &0.58  &0.22  &0.16 \\ 
ASPP  &0.52 &0.17 &0.12 &0.53  &0.18 &0.13 \\
SPAP(fine-to-coarse)  &0.71 &0.20  &0.15  &0.66  &0.20 &0.16 \\
SPAP(coarse-to-fine)  & \textbf{0.73} & \textbf{0.22} & \textbf{0.17}   &\textbf{0.71} &\textbf{0.25} &\textbf{0.19} \\
\hline
SCAN  & 0.64  & 0.20   &0.16 & 0.72  & 0.25   &0.20  \\
Pix2Pix &0.71  &0.25  &0.18 &0.85  &0.40  &0.32  \\
\hline
\end{tabular}
\end{table*}

\begin{figure*}
	\centering
	\begin{tabular}{ccccc}
	\includegraphics[width = 1.0in]{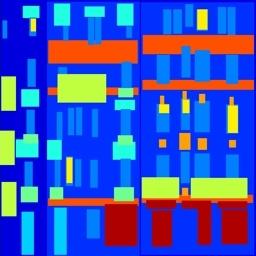} &
	\includegraphics[width = 1.0in]{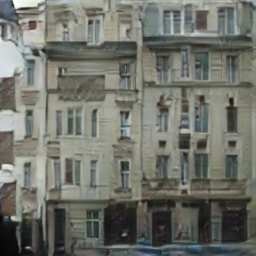} &
	\includegraphics[width = 1.0in]{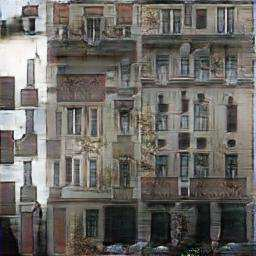} &
	\includegraphics[width = 1.0in]{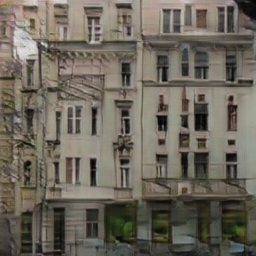} &
	\includegraphics[width = 1.0in]{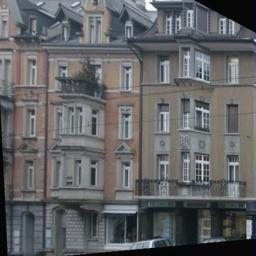} \\
	
	\includegraphics[width = 1.0in]{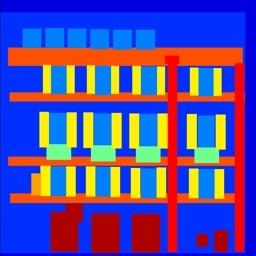} &
	\includegraphics[width = 1.0in]{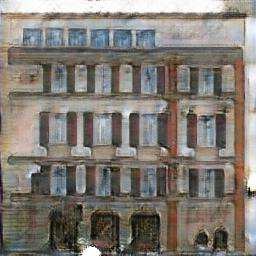} &
	\includegraphics[width = 1.0in]{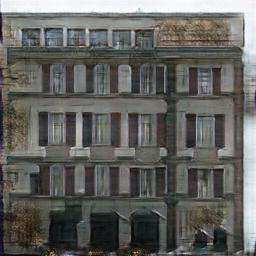} &
	\includegraphics[width = 1.0in]{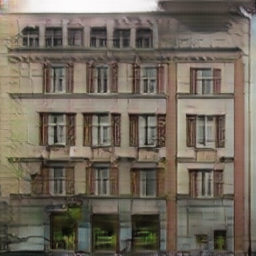} &
	\includegraphics[width = 1.0in]{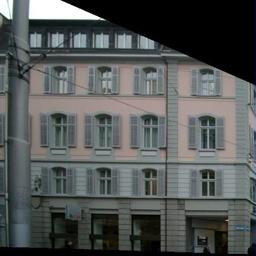} \\
	
	\includegraphics[width = 1.0in]{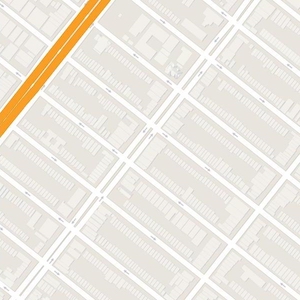} &
	\includegraphics[width = 1.0in]{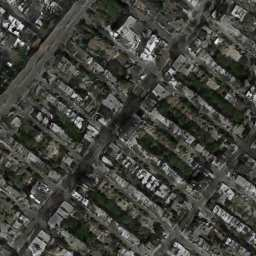} &
	\includegraphics[width = 1.0in]{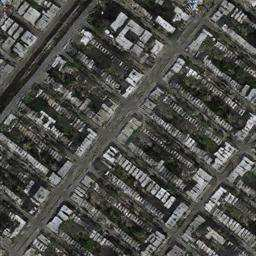} &
	\includegraphics[width = 1.0in]{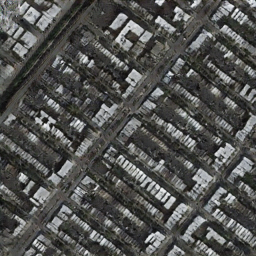} &
	\includegraphics[width = 1.0in]{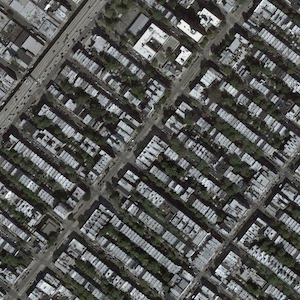} \\
	
	\includegraphics[width = 1.0in]{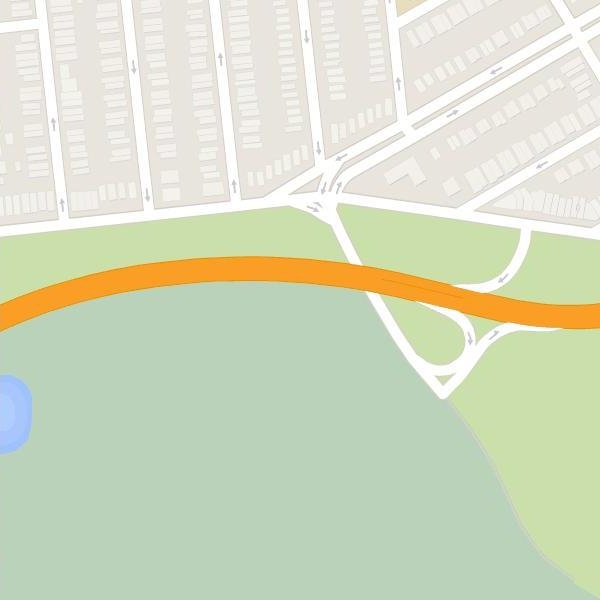} &
	\includegraphics[width = 1.0in]{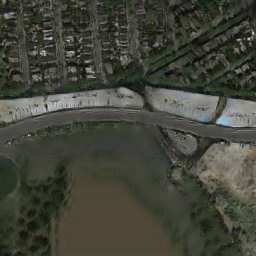} &
	\includegraphics[width = 1.0in]{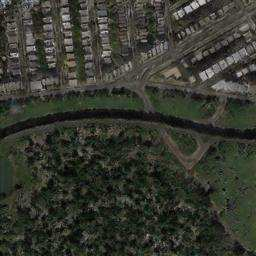} &
	\includegraphics[width = 1.0in]{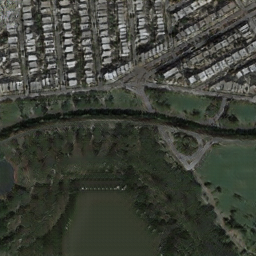} &
	\includegraphics[width = 1.0in]{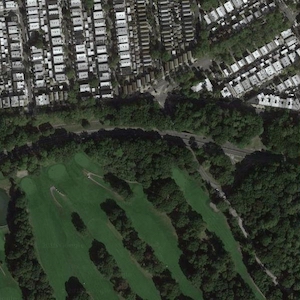} \\
	
	Input & CycleGAN & SCAN & Ours & Ground Truth \\
	\end{tabular}
	\vspace{-1mm}
	\caption{Results on Labels $\Rightarrow$ Facades and Labels $\Rightarrow$ Maps. CycleGAN and SCAN images from paper \cite{Li_2018_ECCV}. Our results generated with SPAP in coarse-to-fine order.} 
	\vspace{-1mm}
	\label{fig:facades}
\end{figure*}

\begin{table*}
	\caption{PSNR and SSIM values Map $\Leftrightarrow$ Aerial and Facades $\Leftrightarrow$ Labels. Note that SCAN has much bigger models since it uses stacked CycleGANs without sharing parameters}  
	\label{maps-ssim}
	\centering
	\begin{tabular}{ccccccccc}
	\hline
	\multirow{2}{*}{Method}  &\multicolumn{2}{c}{\textbf{Aerial $\Rightarrow$ Map}}  & \multicolumn{2}{c}{\textbf{Map $\Rightarrow$ Aerial}}  &\multicolumn{2}{c}{\textbf{Facades $\Rightarrow$ Labels}}  & \multicolumn{2}{c}{\textbf{Labels $\Rightarrow$ Facades}} \\
	 & \textbf{PSNR} &\textbf{SSIM} &\textbf{PSNR} &\textbf{SSIM}  & \textbf{PSNR} &\textbf{SSIM} &\textbf{PSNR} &\textbf{SSIM}\\
	\hline
	CycleGAN  &24.68 & 0.63 & 14.39 &0.20 &8.73 &0.33 &11.72 &0.20 \\
	ASPP  &24.33 &0.65 &\textbf{14.62} &0.21  &7.69 &0.26 &11.65 &0.14 \\
	SPAP(fine-to-coarse)  &24.95 &0.66  &14.47 &0.21  &9.00 &0.35  &11.65 & 0.19\\
	SPAP(coarse-to-fine)  &\textbf{25.02} &\textbf{0.66}  &14.40  &\textbf{0.22} &\textbf{9.21} &\textbf{0.35}  &\textbf{12.20}  &\textbf{0.21}\\
	\hline
	SCAN  &25.15 &0.67 &14.93 &0.23  &8.28 &0.29 &10.67 &0.17 \\
	\hline
	\end{tabular}
\end{table*}

\begin{figure*}
	\centering
	\begin{tabular}{cccccc}

	\includegraphics[width = 1.0in]{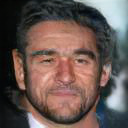} &
	\includegraphics[width = 1.0in]{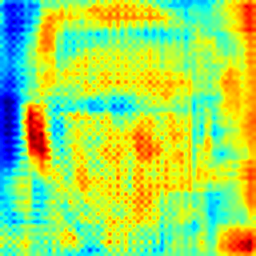} &
	\includegraphics[width = 1.0in]{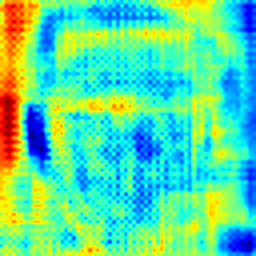} &
	\includegraphics[width = 1.0in]{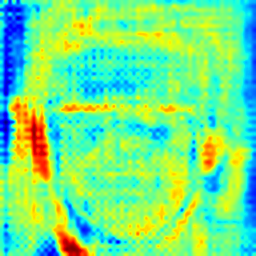} &
	\includegraphics[width = 1.0in]{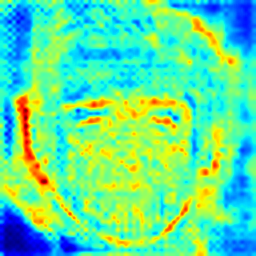} &
	\includegraphics[width = 1.0in]{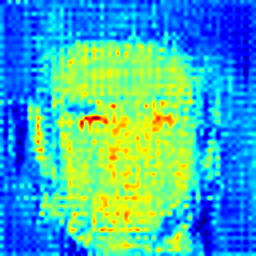}\\
	
	\includegraphics[width = 1.0in]{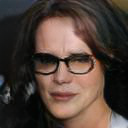} &
	\includegraphics[width = 1.0in]{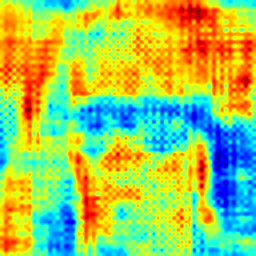} &
	\includegraphics[width = 1.0in]{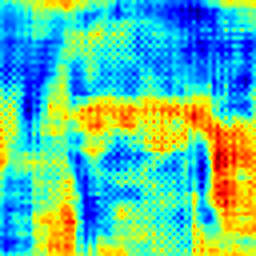} &
	\includegraphics[width = 1.0in]{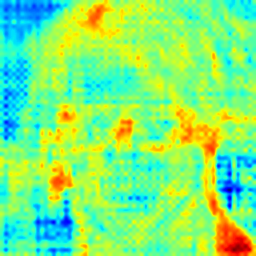} &
	\includegraphics[width = 1.0in]{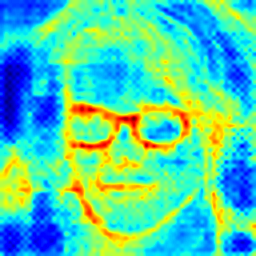} &
	\includegraphics[width = 1.0in]{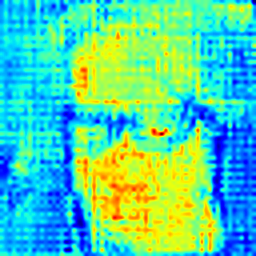}\\
	
	\includegraphics[width = 1.0in]{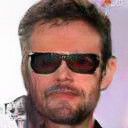} &
	\includegraphics[width = 1.0in]{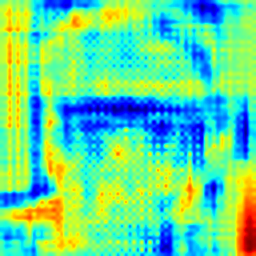} &
	\includegraphics[width = 1.0in]{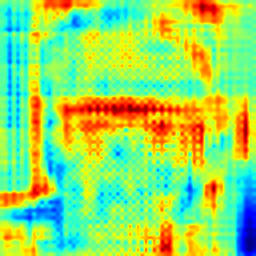} &
	\includegraphics[width = 1.0in]{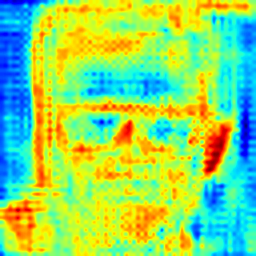} &
	\includegraphics[width = 1.0in]{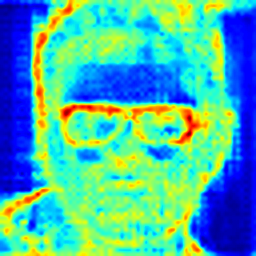} &
	\includegraphics[width = 1.0in]{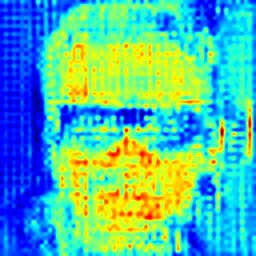}\\

	Synthesized Image & C3D7 & C3D5 & C3D3 & C3D1 & C1D1 \\
	\end{tabular}
	\vspace{-1mm}
	\caption{Selected samples of synthesized images and visualization of attention map for SPAP in coarse-to-fine order. C$k$D$n$ denotes k$\times$k Convolution layer with atrous rate $n$.} 
	\label{fig:gan_coarse_to_fine_vis}
\end{figure*}

\begin{figure*}
\centering
\begin{tabular}{ccccccc}

\includegraphics[width = 0.8in]{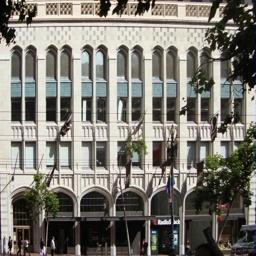} &
\includegraphics[width = 0.8in]{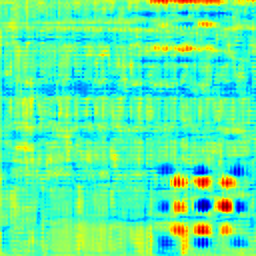} &
\includegraphics[width = 0.8in]{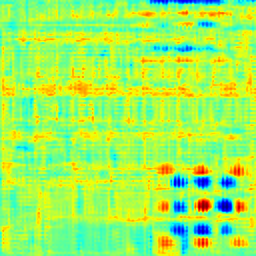} &
\includegraphics[width = 0.8in]{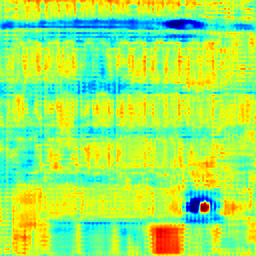} &
\includegraphics[width = 0.8in]{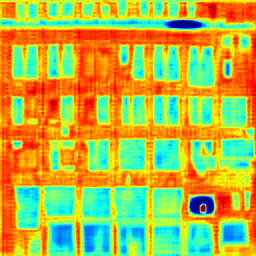} &
\includegraphics[width = 0.8in]{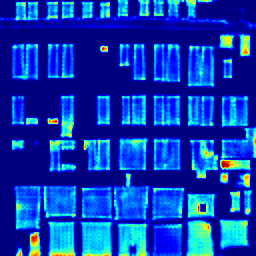} &
\includegraphics[width = 0.8in]{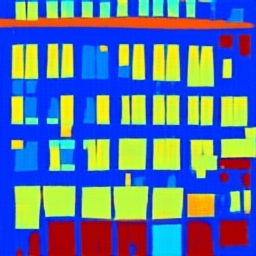}\\

\includegraphics[width = 0.8in]{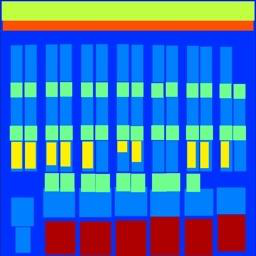} &
\includegraphics[width = 0.8in]{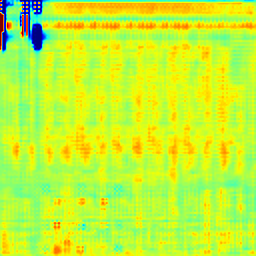} &
\includegraphics[width = 0.8in]{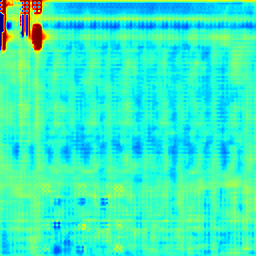} &
\includegraphics[width = 0.8in]{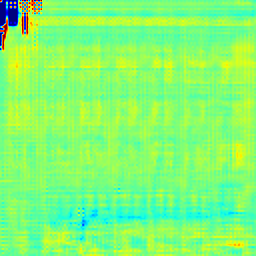} &
\includegraphics[width = 0.8in]{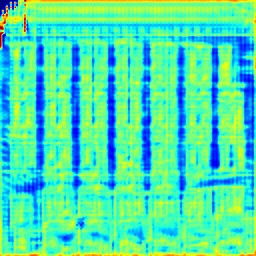} &
\includegraphics[width = 0.8in]{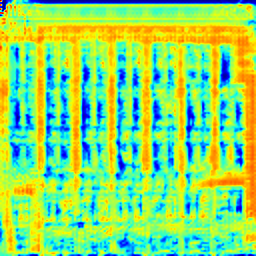} &
\includegraphics[width = 0.8in]{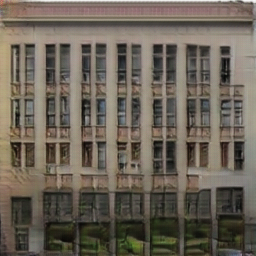}\\

Inputs &  C3D18 &  C3D12 &  C3D6 &  C3D1 &  C1D1 & Outputs\\
\end{tabular}
\vspace{-1mm}
\caption{Selected samples of  synthesized images and visualization of attention mapS for SPAP in coarse-to-fine order.}
\label{fig:cyclegan_atten_vis}
\end{figure*}

\begin{figure*}[!h]
\centering
\begin{tabular}{cccccc}

\includegraphics[width = 1.0in]{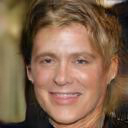} &
\includegraphics[width = 1.0in]{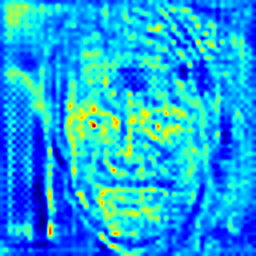} &
\includegraphics[width = 1.0in]{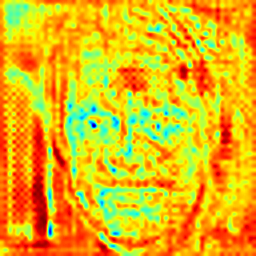} &
\includegraphics[width = 1.0in]{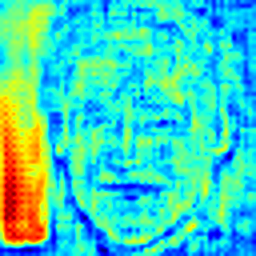} &
\includegraphics[width = 1.0in]{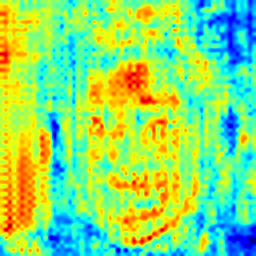} &
\includegraphics[width = 1.0in]{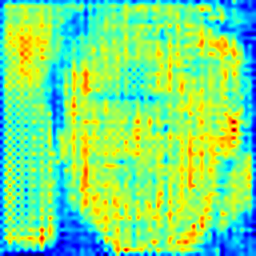} \\

\includegraphics[width = 1.0in]{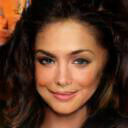} &
\includegraphics[width = 1.0in]{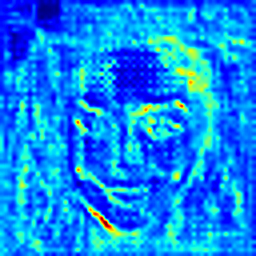} &
\includegraphics[width = 1.0in]{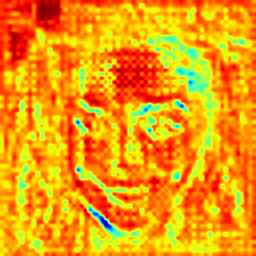} &
\includegraphics[width = 1.0in]{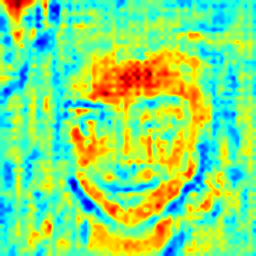} &
\includegraphics[width = 1.0in]{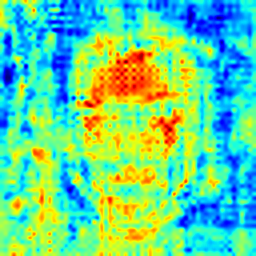} &
\includegraphics[width = 1.0in]{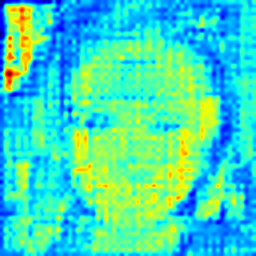} \\

Synthesized Image & C1D1 & C3D1 &  C3D3 & C3D5 & C3D7 \\
\end{tabular}
\vspace{-1.5mm}
\caption{Selected samples of  synthesized images and visualization of attention map for SPAP in fine-to-coarse order.}
\label{fig:gan_fine_to_coarse_vis}
\end{figure*}

\textbf{Image-to-Image translation with CycleGAN} Fig.~\ref{fig:cityscapes} visually compare our results with CycleGAN, SCAN and Pix2Pix. Images generated by SPAP in coarse-to-fine order are more relastic and vivid, and closer to Pix2Pix and ground truth when compared with CycleGAN. This quality improvement are further illustrated in Table.~\ref{cityscape_fcn}, where our model outperforms SCAN and CycleGAN on Label $\Rightarrow$ Image generation, even close to Pix2Pix which is trained on paried data. Our model also outperform CycleGAN on Photo $\Rightarrow$ Label and gets comparable results with SCAN.  

Fig.~\ref{fig:facades} shows selected generated sample results in the Aerial $\Rightarrow$ Map and the Labels $\Rightarrow$ Facades task. We can observe that our results contains better image quality and finer pattern.
 Table.~\ref{maps-ssim} showes quatitive evaluation metric PSNR and SSIM on Map $\Leftrightarrow$ Aerial and Facades $\Leftrightarrow$ Labels dataset, which proves that our image synthesis results are more similar to ground truth in terms of colors and structures than CycleGAN, and results are comparable to SCAN.

\subsection{Analysis of Attention module in Generator $\mathbb{G}$}
Fig.~\ref{fig:gan_coarse_to_fine_vis} and ~\ref{fig:cyclegan_atten_vis} illustrates selected synthesized samples from GAN with SPAP module and visualization of attention map when fusing different feature maps in coarse-to-fine order, and Fig.~\ref{fig:gan_fine_to_coarse_vis} shows attention maps in fine-to-coarse order SPAP. These figures can clearly demonstrate effectiveness of spatial attention in SPAP to fuse different scale information. In face generation in Fig.~\ref{fig:gan_coarse_to_fine_vis}, where SPAP is fused in a coarse-to-fine order, for features of large atrous rates, the model tends to focus more on region like background, hair, etc., while for features of small atrous rates, attention value is high on contour of face, glasses, eyes, etc.


\subsection{Analysis of Receptive field in Discriminator $\mathbb{D}$}
For CycleGAN and Pix2Pix \cite{isola2017image}, PatchGAN is used as discriminator, with a 70x70 receptive field. The PatchGAN only penalizes structure at the scale of patches, it tries to classify if each NxN patch in an image is real or fake. Such a discriminator effectively models the image as a Markov random field, assuming independence between pixels separated by more than a patch diameter. In this paper, we add three parallel convolution with different atrous rates, which will increase the receptive field of the PatchGAN. If we add stride-2 convolution with atrous rate 3,5,7 at 64x64 feature map, the receptive field of discriminator will increase to 236x236. In \cite{isola2017image}, full 286x286 ImageGAN produce a lower FCN-score which, to their opinion, might because that ImageGAN having more parameters and greater depth make it harder to train. In our model, the added atrous convolution can effectively increase the receptive field without adding too many parameters or increasing depth of the model. 

In vanilla SNDCGAN, the receptive field of the convolution before the final fully connected layer is 52. If we add SPAP layer to the model, the receptive field will be increased to 100 by stride-2 convolution with atrous rate 7 in 64x64 feature map. 


\section{Conclusions}
This paper proposed an spatial pyramid attentive pooling (SPAP) building block which can be integrated into both generators and discriminators in GANs based generative learning. The proposed SPAP is a simple yet effective  module which harnesses the advantages of three ubiquitous components in neural architecture design in a novel way: Atrous spatial pyramid for capturing multi-scale information, a cascade attention scheme for aggregating information between differnt levels in the pyramid, and residual connections. The proposed SPAP module is complementary to many existing efforts towards building more accurate and more robust GANs based generative learning models. In experiments, we test our method on unconditional GANs with the Celeba-HQ-128 dataset and unpaired image-to-image CycleGANs with the CityScape, Facade, Maps dataset, all obtaining better or comparable performance than state-of-the-art models. 

\noindent\textbf{Acknowledgment.} This work is supported by ARO award W911NF1810295, ARO DURIP award W911NF1810209 and NSF IIS 1822477.

\bibliographystyle{plain}
\bibliography{egbib}

\end{document}